\documentclass[letterpaper, 10 pt, conference]{ieeeconf}  %
\usepackage{amssymb}
\usepackage{comment}
\usepackage{myconfig}

\IEEEoverridecommandlockouts                        

\title{\LARGE \bf
PHASE: Compliance-Enabled Tactile Phase Retrieval \\for Few-Shot Insertion Learning
}

\author{%
Jeremy Siburian$^{1*,2}$%
\quad Cristian C. Beltran-Hernandez$^{1}$%
\quad Tatsuya Matsushima$^{2}$\\
Yusuke Iwasawa$^{2}$%
\quad Masashi Hamaya$^{1\dagger}$%
\quad Mai Nishimura$^{1\dagger}$%
\thanks{$^{1}$ OMRON SINIC X Corporation, 5-24-5 Hongo, Bunkyo-ku, Tokyo, 113-0033, Japan (e-mail: {\tt\small masashi.hamaya@sinicx.com}).}
\thanks{$^{2}$ The University of Tokyo, 7-3-1 Hongo, Bunkyo-ku, Tokyo, 113-8654, Japan.}
\thanks{$^{*}$ Work done during internships at OMRON SINIC X.}
\thanks{$^{\dagger}$ Joint last authors.}
\thanks{This work was supported by JST, Japan PRESTO Grant Number JPMJPR2518 and  JPMJPR24T9.}
}

\begin{document}

\twocolumn[{%
\noindent
\copyright{} 2026 IEEE. Personal use of this material is permitted. Permission from IEEE must be obtained for all other uses, in any current or future media, including reprinting/republishing this material for advertising or promotional purposes, creating new collective works, for resale or redistribution to servers or lists, or reuse of any copyrighted component of this work in other works.\\

\noindent
\textbf{Accepted article:}\\
J. Siburian, C. C. Beltran-Hernandez, T. Matsushima, Y. Iwasawa, M. Hamaya, and M. Nishimura, ``PHASE: Compliance-Enabled Tactile Phase Retrieval for Few-Shot Insertion Learning,'' \textit{IEEE/RSJ International Conference on Intelligent Robots and Systems}, 2026.
}]
\thispagestyle{empty}
\pagenumbering{gobble}
\clearpage

\maketitle
\thispagestyle{empty}
\pagestyle{empty}

\begin{abstract}
Contact-rich assembly tasks such as peg-in-hole insertion remain difficult to learn from limited demonstrations. While retrieval-augmented imitation learning, which augments target demonstrations with relevant prior data, offers a promising direction, its applicability to contact-rich manipulation remains largely unexplored. Contact-rich insertion unfolds over multiple phases from \emph{search} to \emph{insert}, and retrieving phase-specific experience from prior data in principled ways remains an open question. Our key insight is that a compliant wrist enables the robot to sustain contact throughout execution, producing rich tactile and force signals that naturally reveal the phase structure of insertion and inform what should be retrieved. Based on this insight, we present \textbf{PHASE} (\textbf{PH}ase-\textbf{A}ware \textbf{S}egmentation and R\textbf{E}trieval), a framework for compliance-enabled tactile phase retrieval that integrates multimodal contact-aware representation learning, variable-length phase segmentation from tactile signals, and phase-consistent retrieval for policy learning. We evaluate PHASE on real-world peg-in-hole insertion across five peg geometries, comparing against retrieval strategies drawn from state-of-the-art methods under a shared policy architecture. 
PHASE improves the overall success rate by 13 percentage points over the strongest non-phase-aware baseline, and improves performance under unseen initial positions by 30 percentage points. These results demonstrate that aligning retrieval with interaction-defined contact phases substantially improves robustness in few-shot insertion learning.
\end{abstract}

\section{Introduction}
\label{sec:introduction}

A long-standing goal in robot manipulation is to enable robots to acquire new skills from limited demonstrations by reusing prior task experience. Neural and cognitive studies show that humans naturally do this: by relying on memory and remembering past experiences, humans are able to generalize quickly when trying to learn a new task~\cite{tulving2002episodic,shadmehr1997neural, hall2025neural}. 
For example, someone who has learned to insert a few types of connectors can draw on memory and prior experience to handle a new connector shape, leveraging sensorimotor memory from similar insertions and adapting to the new geometry.
This brings us to the question:
\textit{how can we enable robots to leverage prior experience in the same way?}

In this work, we address this question in the context of peg-in-hole insertion: given a few demonstrations
of a new peg shape, we retrieve relevant segments from a library of prior insertion experience to enable generalization across geometries.
To this end, we adopt retrieval-based imitation learning, a promising approach that, given a small set of target-task demonstrations, searches a larger database of prior experience for relevant segments and combines them with the target data to train a policy~\cite{nasiriany2022sailor, humphreys2022largescale, di2024effectiveness, kuang2025ram}. While these methods have shown strong results on tasks like pick-and-place~\cite{du2023behavior, lin2024flowretrieval}, contact-rich tasks remain underexplored. Peg-in-hole insertion, in particular, requires the robot to navigate tight geometric tolerances through
force-driven interaction: searching for a hole, aligning under contact, and executing a controlled insertion. Even small distribution shifts can cause policies trained on limited demonstrations to fail, making it a compelling setting for retrieval-based learning.

\begin{figure}[t]
    \centering
    \includegraphics[width=\linewidth]{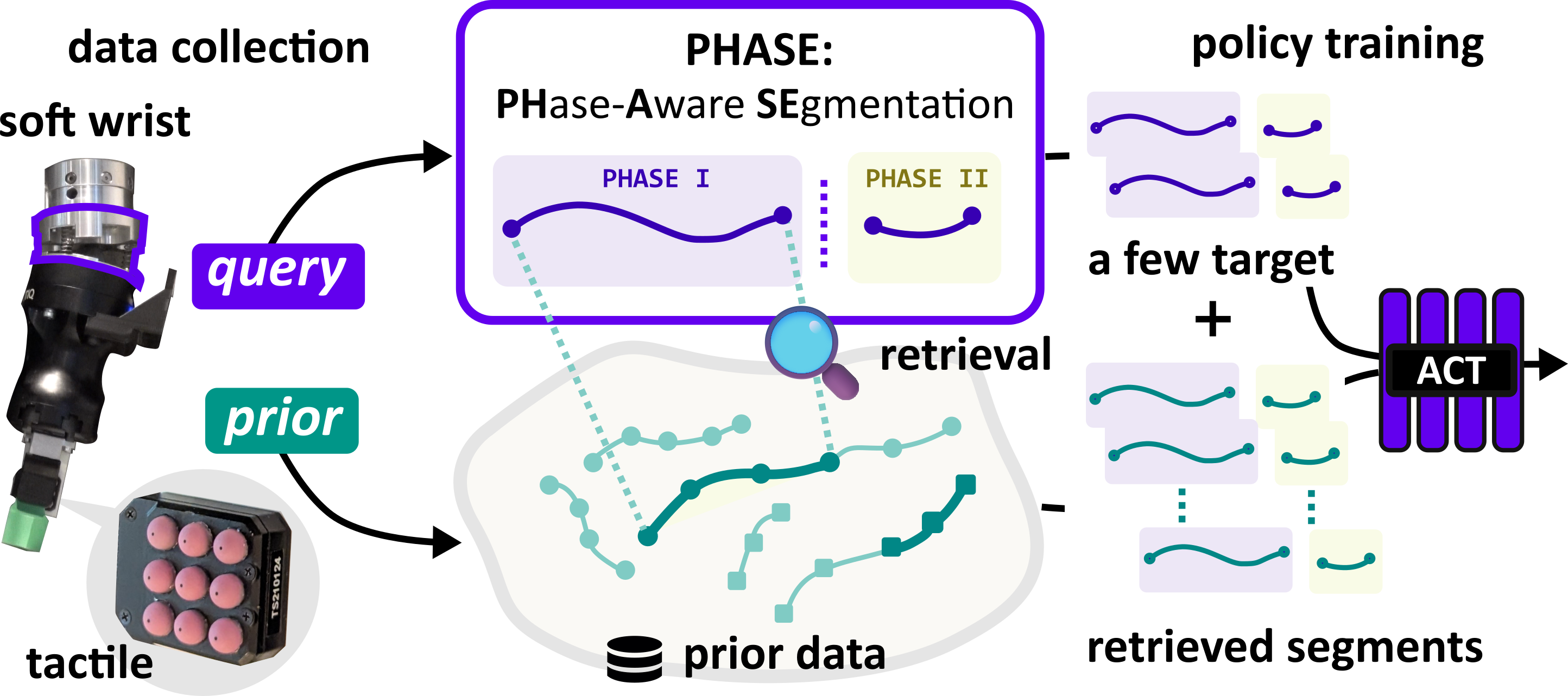}
    \caption{\textbf{PHASE: Phase-Aware Segmentation \& Retrieval.} Given a few target task demonstrations, PHASE leverages compliance-enabled tactile and force signals to segment prior demonstrations into variable-length contact phases, retrieves phase-consistent segments, and trains robust policies that generalize across geometries and distribution shifts.}
    \label{fig:teaser}
    \figuretotextspace
\end{figure}

Applying retrieval to insertion tasks introduces several challenges. First, the boundaries between meaningful task phases, such as the transition from searching to inserting, are reflected in force and tactile signals rather than kinematic changes. Second, the duration of each phase varies across demonstrations depending on geometry and initial conditions. As a result, retrieval based on kinematic similarity may combine fragments of different interaction regimes or split coherent contact behaviors across retrieval boundaries, degrading the training signal \cite{strap2025, belkhale2024rt, myers2024policy}. Cognitive research shows that humans naturally segment continuous activity at salient sensory boundaries and selectively retrieve relevant fragments, and that identifying appropriate boundaries leads to better recall and more effective learning~\cite{zacks2007event}. This suggests that grounding retrieval in interaction signals should similarly improve policy learning, yet prior work has not pursued this direction, in part because it requires interaction signals that are rich and informative throughout execution.

Compliance and tactile sensing provide the key to addressing these challenges. The key insight is that \textit{a compliant wrist absorbs contact forces and prevents protective stops, enabling the robot to sustain contact throughout execution and produce rich tactile and force signals that naturally reveal the phase structure of insertion and inform what should be retrieved}. On a rigid manipulator, contact forces can quickly exceed safety thresholds, triggering protective stops before the full interaction sequence can be observed~\cite{morgan2023towards}. Even when execution completes, misalignments produce large forces that jam the peg, yielding signals dominated by impact spikes rather than the gradual transitions that define contact phases. With compliance, the resulting signals expose when interaction events occur and how long each contact phase persists, providing a natural decomposition of insertion behavior. Effective retrieval should therefore align with this interaction-defined structure rather than impose uniform temporal segmentation.

Based on this insight, we propose \textbf{PHASE} (\underline{\textbf{PH}}ase-\underline{\textbf{A}}ware \underline{\textbf{S}}egmentation and R\underline{\textbf{E}}trieval), a framework for \textit{compliance-enabled tactile phase retrieval} that structures retrieval around interaction-defined task phases revealed by tactile and force signals.
\Cref{fig:teaser} illustrates an overview of our framework.
To collect both target and prior demonstrations, we use the compliant robot setup, which comprises a soft wrist~\cite{von2020compact} and a distributed tactile sensor~\cite{khamis2019novel} that captures rich contact information throughout execution.
We first construct a prior database by encoding prior demonstrations via Masked Tactile Trajectory Transformer (MAT$^3$)~\cite{kamijo2026tactile}.
Given a few target demonstrations, we segment trajectories 
into variable-length contact phases by detecting phase boundaries from tactile signals, yielding segments aligned with interaction events such as contact initiation and insertion onset. 
Finally, to train policies for new tasks, we retrieve phase-consistent segments from prior data and combine them with the target demonstrations to train a policy
~\cite{zhao2023learning}.
We evaluate PHASE on real-world peg-in-hole insertion across five peg geometries (two seen, three unseen)
and under unseen starting positions. 
PHASE improves the overall success rate by 13 percentage points over the strongest non-phase-aware baseline and by 30 percentage points under state distribution shift.

Our contributions are summarized as follows:
\begin{itemize}
\item \textbf{Compliance-Enabled Tactile Phase Retrieval}: We show that
  aligning retrieval with contact phases revealed by
  compliance-enabled tactile signals substantially
  improves robustness in few-shot insertion
  learning, with a 30-percentage-point improvement under state distribution shift over the strongest non-phase-aware baseline.
\item \textbf{PHASE Framework}: To realize this, we propose PHASE, a retrieval framework that leverages compliance-enabled tactile and force signals to structure retrieval around interaction-defined contact phases. PHASE integrates multimodal contact-aware representation learning, variable-length phase segmentation from tactile signals, and phase-consistent retrieval for policy learning.
\end{itemize}

\section{Related Work}
\label{sec:related_work}

\begin{figure*}[t]
    \centering
    \includegraphics[width=\linewidth]{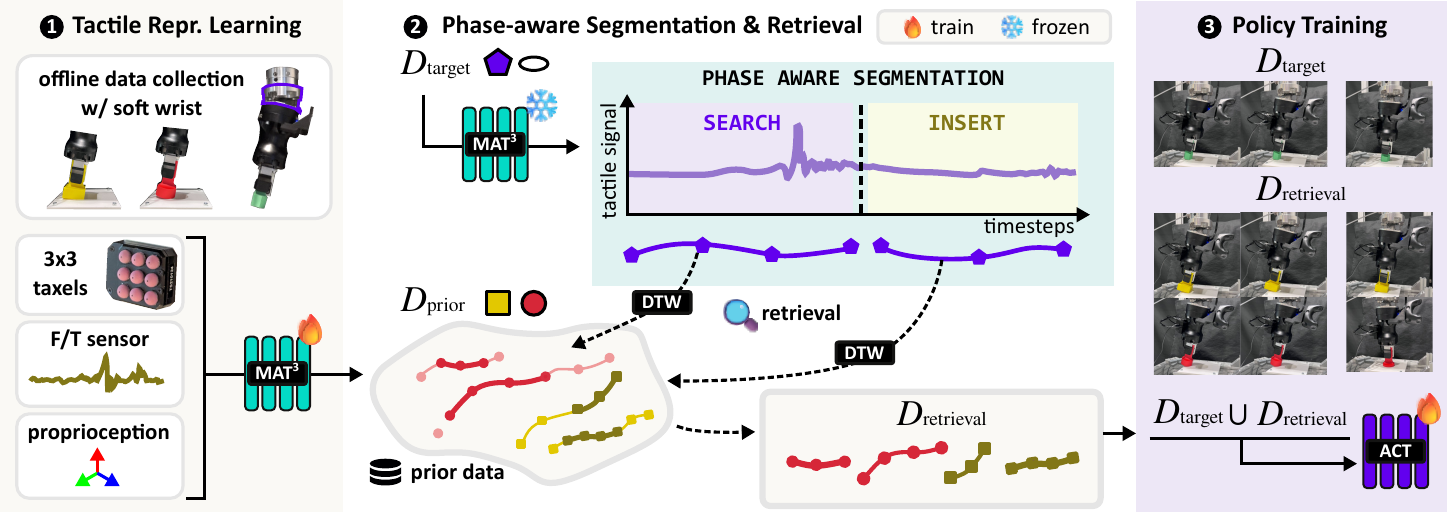}
    \caption{\textbf{Overview of PHASE.} The framework
    consists of three stages:
    (1) \textbf{Tactile Representation Learning}, where
    a masked tactile-proprioceptive
    encoder~\cite{kamijo2026tactile} learns embeddings
    that capture force-driven interaction dynamics;
    (2) \textbf{Phase-Aware Segmentation \& Retrieval},
    where demonstrations are segmented using
    tactile-estimated torque signals to identify
    interaction-defined contact phases and
    variable-length segments are retrieved accordingly;
    and
    (3) \textbf{Policy Training}, where an ACT policy~\cite{zhao2023learning} is trained on the retrieved, phase-consistent segments for robust \rev{peg-in-hole} insertion.}
    \label{fig:overview}
    \figuretotextspace
\end{figure*}

\subsection{Contact-Rich Manipulation Tasks}
Contact-rich manipulation, such as peg-in-hole insertion, requires precise modeling of contact transitions, alignment strategies, and force modulation \cite{beltran2020variable}. Classical approaches rely on impedance control and hybrid force-position control to regulate interaction forces during insertion \cite{hogan1984impedance, raibert1981hybrid}. More recently, learning-based methods have sought to acquire insertion skills from demonstrations using behavior cloning \cite{ankile2024juicer, wang2022adaptive} or reinforcement learning \cite{schoettler2020deep}. 
Despite these advances, learning contact-rich behaviors from limited data remains challenging. Insertion involves discontinuous contact transitions and geometry-dependent interaction strategies that are highly sensitive to initial conditions. Small variations in geometry or pose can induce qualitatively different contact regimes, making generalization difficult. 
Prior work has shown that policies trained across diverse assembly tasks often degrade under geometric variation due to negative transfer and conflicting interaction dynamics~\cite{tang2023industreal, guo2025srsa}.
We address this by structuring demonstration reuse around interaction-level contact phases revealed by compliance-enabled tactile signals.

\subsection{Retrieval-Based Imitation Learning}
Retrieval-based imitation learning has emerged as an alternative to fully parametric multi-task models~\cite{humphreys2022largescale}. Rather than encoding all prior experience in a single network,
these methods retrieve relevant demonstrations from an offline corpus to guide adaptation to new tasks.
Behavior Retrieval selects state-action pairs for few-shot imitation learning~\cite{du2023behavior},
while FlowRetrieval measures motion similarity via optical flow representations~\cite{lin2024flowretrieval}. Other works retrieve skills or affordances through learned embeddings to enable cross-task
transfer~\cite{nasiriany2022sailor, kuang2025ram}. More recently, sub-trajectory-based approaches
decompose demonstrations into reusable segments~\cite{belkhale2024rt}. STRAP retrieves variable-length sub-trajectories using temporal alignment~\cite{strap2025}, showing strong performance
on multi-stage pick-and-place tasks. However, these methods are evaluated primarily on motion-dominant manipulation and do not address contact-rich settings, where task progression is governed by interaction
forces.
A separate line of work has addressed this by grounding retrieval in tactile and force signals,
matching prior trajectories using force or tactile similarity~\cite{kamijo2026tactile, yu2024mimictouch,
pmlr-v229-guzey23a}. These approaches typically perform non-parametric retrieval at execution time,
functioning as memory-based controllers. While effective when closely matching demonstrations exist,
they rely on direct trajectory matching rather than training parametric policies, limiting generalization
beyond stored experiences. SRSA addresses contact-rich assembly through policy-level retrieval, selecting and
adapting previously trained policies via reinforcement learning~\cite{guo2025srsa}. However, retrieval
operates at the policy level rather than structuring demonstration data for imitation learning. 
In summary, how retrieval should be structured to
support few-shot \rev{parametric} policy learning in
contact-rich settings remains an open question.
We address this by retrieving phase-consistent, variable-length segments grounded in compliance-enabled interaction signals.

\subsection{Tactile and Contact Phase Representations}
Tactile sensing and force/torque feedback are widely used for contact detection, insertion stabilization, and reactive control \cite{dong2021tactile, jin2024visual}. 
Heuristic approaches estimate contact states from tactile signals to achieve stable grasps~\cite{romano2011human, jang2025tactile}, and force/torque sensors have been used for contact state estimation in peg-in-hole tasks~\cite{morgan2023towards, lee2022contact}. Recent work shows that incorporating torque signals into policy architectures improves contact-rich manipulation by encouraging physically grounded representations of interaction dynamics~\cite{zhang2025ta}.
Building on these signals, several works decompose insertion into primitives or phases to facilitate control or modeling~\cite{zhang2022learning, schoettler2020deep}.
However, phase decompositions are typically used for feedback control rather than to structure retrieval. In contrast, our approach uses compliance-enabled tactile and force signals both to segment demonstrations at interaction-defined phase boundaries and to compute retrieval similarity, grounding the entire retrieval pipeline in contact dynamics.

\section{Methodology}
\label{sec:method}

Figure~\ref{fig:overview} summarizes PHASE's three stages: tactile representation learning, phase-aware segmentation and retrieval, and ACT policy training.

\subsection{Problem Formulation}
\label{subsec:problem}
We consider the problem of learning peg-in-hole
insertion from limited demonstrations in a few-shot
imitation learning setting. We focus on real-world
insertion using a rigid robotic arm equipped with a
soft wrist and force-torque
sensing~\cite{von2020compact}.
Each demonstration is a trajectory $\tau=\{(s_t, a_t)\}_{t=1}^T$, where $s_t$ denotes the robot state and $a_t$ the corresponding action at timestep $t$. The state $s_t$ includes tactile readings, wrist force--torque measurements, and robot proprioceptive signals (joint positions and end-effector pose), while the action $a_t$ corresponds to the commanded end-effector pose.
We assume access to a prior dataset
$\mathcal{D}_{\text{prior}}$ consisting of
demonstrations from related insertion tasks (e.g.,
different peg geometries), and a small target dataset
$\mathcal{D}_{\text{target}}$ containing only a few
demonstrations of the task of interest. Our goal is to
retrieve phase-consistent segments from
$\mathcal{D}_{\text{prior}}$ and combine them with
$\mathcal{D}_{\text{target}}$ to form an augmented
training set
$\mathcal{D}_{\text{train}} =
\mathcal{D}_{\text{target}} \cup
\mathcal{D}_{\text{ret}}$.
We then \rev{train a parametric policy}
$\pi(a \mid s)$ via behavior
cloning (BC) on $\mathcal{D}_{\text{train}}$, 
enabling few-shot adaptation to target geometries absent from the prior dataset and generalization to unseen starting positions through retrieval-based data augmentation.

\subsection{Tactile Representation Learning}
\label{sec:representation_learning}
Retrieval in peg-in-hole insertion requires embeddings
that capture interaction dynamics and contact-phase
transitions, so that similarity in embedding space
reflects similarity in contact behavior. To this end,
we adopt the Masked Tactile Trajectory Transformer
(MAT$^3$)~\cite{kamijo2026tactile} to learn multimodal
tactile-proprioceptive representations. MAT$^3$ learns
per-frame embeddings through self-supervised masked
reconstruction, requiring no task-specific labels or
phase annotations. Each per-frame embedding
$\bm z_t \in \mathbb{R}^{256}$ is contextualized by a
temporal history window via bidirectional
attention~\cite{wu2023masked}, encoding not only the
current contact state but also how that state relates
to its temporal neighbors. The masked objective forces
the encoder to jointly reason across tactile patterns,
forces, and proprioception, producing embeddings that
capture cross-modal interaction dynamics.
At each timestep, tactile signals from a $3\times3$
taxel (tactile pixel) grid and the action form base
tokens, while force/torque, arm pose, gripper pose,
and temporal information serve as auxiliary context.
The policy receives the same sensory modalities as the encoder, excluding the expert action used by the offline retrieval encoder, ensuring that segment boundaries and retrieval
distances are defined in the same sensory space the
policy acts upon. For each frame, average pooling over
token embeddings yields
$\bm z_t = g(\mathcal E(\tau)_t)$. A segment of $n$
frames is represented as
$(\bm z_1, \ldots, \bm z_n)$ and matched via DTW.
Unlike~\cite{kamijo2026tactile}, which uses $\bm z_t$
for non-parametric control, we use these per-frame
embeddings for variable-length segment retrieval.
The encoder is trained on
$\mathcal{D}_{\mathrm{prior}}$ for 100 epochs with
history window $T_h{=}15$ and masking ratio sampled from
$[0, 0.6]$.
Aside from these hyperparameters, we use the standard MAT$^3$ architecture and training setup.
The encoder is
frozen thereafter and embeddings are computed offline
before retrieval.

\subsection{Phase-Aware Segmentation \& Retrieval}

\begin{figure}[t]
    \centering
    \includegraphics[width=\columnwidth]{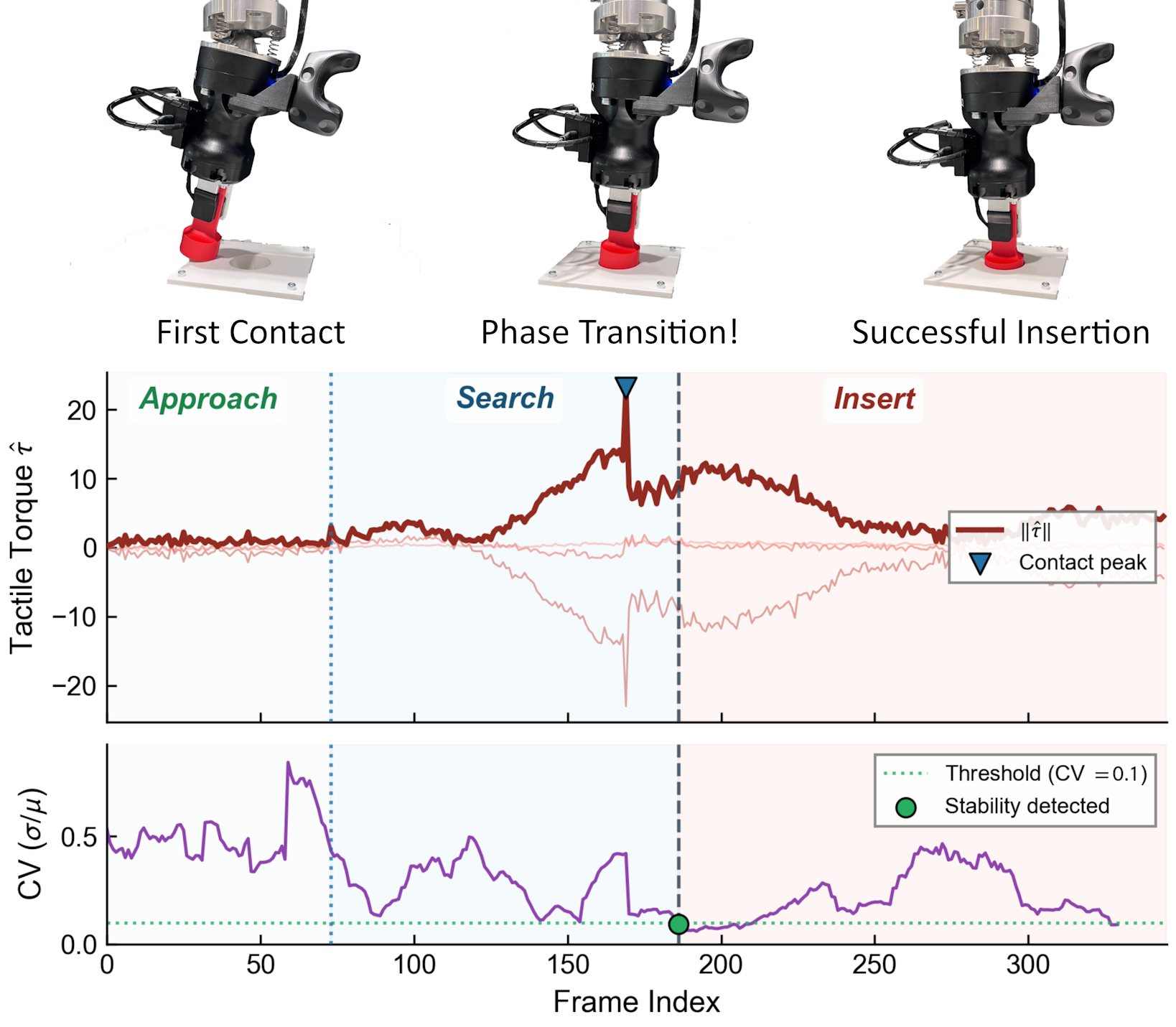}
    \caption{\textbf{Tactile-Based Phase Segmentation.}
    The top panel shows the tactile-estimated torque magnitude $\lVert\hat{\boldsymbol{\tau}}\rVert$ (dark curve) and the signed components of $\hat{\tau}$ (light curves), while the bottom panel shows the coefficient of variation (CV) of the torque magnitude.
    The boundary between \emph{search} and
    \emph{insert} is detected at the first frame where
    $\mathrm{CV} < \eta$ after the contact peak,
    indicating the onset of stable engagement.
    Robot snapshots (inset) show the physical interaction
    during each phase. 
    }
    \label{fig:tactile_segmentation}
    \figuretotextspace
\end{figure}

\textbf{Compliance-Enabled Tactile Phase Segmentation.}
We segment demonstrations at interaction-defined
boundaries rather than at fixed intervals, so that
each retrieved segment corresponds to a coherent
contact phase. Prior segmentation approaches (e.g.,
velocity-based heuristics~\cite{strap2025}) are
effective for motion-dominant tasks but are
insufficient for insertion, where phase transitions are
governed by interaction forces rather than purely
kinematic changes. 
The soft wrist facilitates sustained contact and provides smooth interaction signals that are suitable for the proposed offline phase segmentation.
In contrast, with a rigid wrist, slight contacts can induce large force/torque spikes, making phase discrimination less reliable.
Inspired by~\cite{zhang2025ta}, we detect insertion phase transitions from tactile signals by monitoring torque estimated from distributed tactile readings, $\hat{\boldsymbol{\tau}}=\sum_i \mathbf{r}_i \times \mathbf{f}_i$, where $\mathbf{f}_i$ and $\mathbf{r}_i$ are the 3D force and position of taxel $i$, respectively. In our implementation, taxel positions are modeled on a centered $3 \times 3$ planar grid as $\mathbf{r}_i = s [u_i, v_i, 0]^\top$, with $(u_i, v_i) \in \{-1,0,1\}^2$ and inter-taxel spacing $s = 5\,\mathrm{mm}$. We use tactile-estimated torque rather than wrist-mounted force/torque measurements because it captures the local load distribution at the gripper--peg interface induced by peg--hole contact and is consistent with the rest of the pipeline, where encodings and policy observations are also derived from tactile feedback. Torque is also more discriminative than force magnitude for distinguishing search from insert, as the torque profile changes markedly when the peg aligns with the hole.
We decompose insertion into two phases: (1)~a \emph{search} phase, during which the robot slides to locate the hole, and (2)~an \emph{insert} phase, during which the peg is aligned and pushed downward (Fig.~\ref{fig:tactile_segmentation}). While this two-phase decomposition is specific to insertion, the principle of segmenting at interaction-defined boundaries generalizes to other contact-rich tasks. Given a trajectory, we detect a boundary frame $b$ by computing the coefficient of variation $\mathrm{CV}=\sigma/\mu$ of $\|\hat{\boldsymbol{\tau}}\|$ over a sliding window of $W$ frames and identifying the first frame after the contact peak (the global maximum of $\|\hat{\boldsymbol{\tau}}\|$) where $\mathrm{CV}<\eta$, indicating the onset of stable engagement. The trajectory is then partitioned into a search segment $\sigma^s=(\mathbf{z}_1,\ldots,\mathbf{z}_b)$ and an insert segment $\sigma^r=(\mathbf{z}_{b+1},\ldots,\mathbf{z}_T)$, where $\mathbf{z}_t$ denotes the per-frame embedding from Section~\ref{sec:representation_learning}. Because $b$ depends on contact-transient timing, which varies with geometry and initial conditions, segment lengths are naturally variable. In our experiments we set $W=15$ frames ($0.3$\,s at $50$\,Hz) and $\eta=0.1$. 
A grid search over $W\in\{5,10,15,20,25,30\}$ and $\eta\in\{0.05,0.1,0.15,0.2,0.25\}$ showed that, for $W\in[10,20]$ and $\eta\in[0.1,0.2]$, the method identified a boundary satisfying the stability criterion, i.e., a post-peak frame with $CV < \eta$, in 80--98\% of the episodes.
If the stability criterion is not met, the boundary defaults to the end of a bounded post-peak search interval, ensuring all episodes retain a two-phase segmentation.

\textbf{Phase-Aligned Retrieval.}
Because segment lengths vary across demonstrations,
retrieval must support alignment between sequences of
arbitrary duration. We employ Fast Dynamic Time Warping
(FastDTW)~\cite{salvador2007toward} with a radius of 1 and Euclidean distance between per-frame embeddings to match query and demonstration segments in
embedding space. DTW is preferred over distance
computation on pooled embeddings because it preserves
the temporal ordering of contact events within each
segment, enabling alignment between demonstrations
that progress through the same phases at different
rates. For each query segment $\sigma_q$, we retrieve
the top-$k$ demonstration segments with the smallest
DTW distance, restricting matches to the same phase
($\sigma^s \leftrightarrow \sigma^s$,
$\sigma^r \leftrightarrow \sigma^r$). The retrieved set
is defined as
$\mathcal{D}_{\mathrm{ret}} = \bigcup_q
\mathrm{top}\text{-}k(\sigma_q)$, where $k$ is the
retrieval budget, i.e., the number of prior
segments retrieved per query segment per phase.
\jeremyCR{
A sweep over $k \in \{10, 15, 20, 25\}$ showed that frame-level retrieval coverage increased with diminishing returns and saturated at $k{=}20$, the value we adopt.
}

\subsection{Policy Training}
Once the retrieved set $\mathcal{D}_{\text{ret}}$ is
constructed, it is combined with the target
demonstrations to form the augmented training set
$\mathcal{D}_{\text{train}} =
\mathcal{D}_{\text{target}}
\cup \mathcal{D}_{\text{ret}}$. PHASE is agnostic to
the choice of policy architecture and can be paired
with any imitation learning method. In practice, we
adopt Action Chunking with Transformers
(ACT)~\cite{zhao2023learning}, a Transformer-based
behavior cloning policy. The policy receives a
multimodal observation $\mathbf{o}_t$ formed by
concatenating tactile readings, wrist force/torque
measurements, and robot proprioceptive states. ACT
predicts action chunks $\hat{\mathbf{a}}_{t:t+H}$ over
a horizon $H{=}50$ using a VAE-regularized
transformer, trained on $\mathcal{D}_{\text{train}}$
with MSE reconstruction loss and KL regularization
($\beta{=}1$, $\mathbf{z} \in \mathbb{R}^{32}$).
Each retrieved phase segment is treated as an independent trajectory; actions beyond the segment end are padded and masked from the loss, while target demonstrations remain complete trajectories to preserve the search-to-insert transition.
The policy is
trained with AdamW (lr $= 10^{-4}$, batch size 64)
for 100k gradient steps. Because $\mathcal{D}_{\text{ret}}$ contains only interaction-similar and phase-consistent segments, the augmented training set excludes demonstrations from mismatched
contact regimes, reducing distributional mismatch while
preserving sufficient diversity and phase coverage.

\section{Experiments}
\label{sec:experiments}

\noindent

Our experiments address the following questions:
\begin{itemize}
\item \textcolor{blue}{\textbf{RQ1:}} Does
  retrieval-augmented policy learning improve
  performance compared to learning without retrieval?
\item \textcolor{blue}{\textbf{RQ2:}} Does
  phase-structured retrieval outperform retrieval
  methods that do not account for phase structure?
\item \textcolor{blue}{\textbf{RQ3:}} Does
  phase-structured retrieval improve robustness under
  geometry and state distribution shift?
\item \textcolor{blue}{\textbf{RQ4:}} What failure
  modes persist, and what do they reveal about the
  limitations of the approach?
\end{itemize}

\subsection{Experimental Setup}
\label{subsec:experiment_setup}

\begin{figure}[t]
    \centering
    \includegraphics[width=0.6\linewidth]{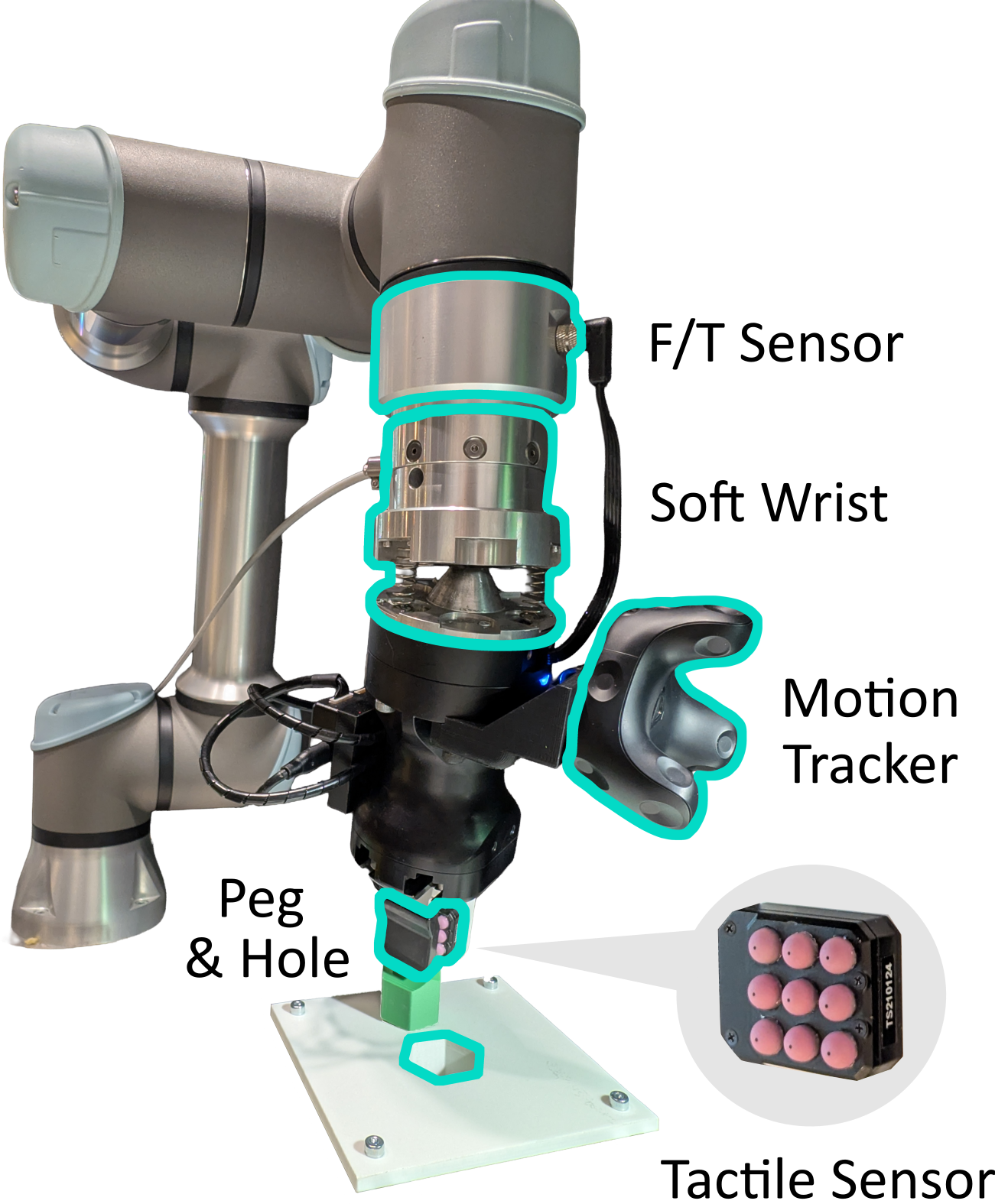}
    \caption{\textbf{Compliant Robot for Tactile Phase Retrieval.}
 The UR5e, which includes an integrated F/T sensor, is equipped with a soft wrist providing passive 6D compliance, and a parallel gripper with a $3\times3$ distributed tactile sensor. Gripper pose is tracked via a motion capture system. Demonstrations are collected through teleoperation for peg-in-hole insertion tasks with multiple peg geometries and varied initial poses.}
    \label{fig:experiment_setup}
    \figuretotextspace
\end{figure}

\textbf{Real Robot Setup.}
Experiments are conducted using a UR5e robotic arm
(Universal Robots A/S, Denmark) equipped with an
integrated force/torque (F/T) sensor, a parallel
gripper (Hand-E, Robotiq, Canada), and a soft wrist
mechanism~\cite{von2020compact}. The soft wrist
consists of three coil springs arranged in parallel,
providing passive 6D compliance that absorbs
interaction forces which would otherwise trigger
protective stops on a rigid manipulator. A distributed
tactile sensor (PapillArray Tactile
Sensor~\cite{khamis2019novel}, Contactile Pty Ltd,
Australia) is mounted on one side of the gripper,
containing a $3\times3$ array of taxels that each
measure 3D force. Gripper pose is obtained via a
motion tracking system (HTC VIVE Tracker, HTC
Corporation, Taiwan). Offline demonstrations are
collected through teleoperation using this setup (Fig.~\ref{fig:experiment_setup}).

\begin{table*}[t]
    \centering
    \footnotesize
    \caption{Real-world success rate over 20 trials per
    shape. Totals are reported over 100 trials per
    method. \hcb{Best} and \hcs{second-best} results
    are highlighted.}
    \label{tab:main_results}
    \begin{tabular*}{\textwidth}{@{\extracolsep{\fill}}lccccc c}
    \toprule
    & \multicolumn{2}{c}{\textbf{Seen Shapes}}
    & \multicolumn{3}{c}{\textbf{Unseen Shapes}}
    & \\
    \cmidrule(lr){2-3} \cmidrule(lr){4-6}
    \textbf{Method}
    & \textbf{Circle} \vspace{2mm}
    & \textbf{Square}
    & \textbf{Rectangle}
    & \textbf{Oval}
    & \textbf{Hexagon}
    & \textbf{Total} \\
    &  \includegraphics[height=1cm]{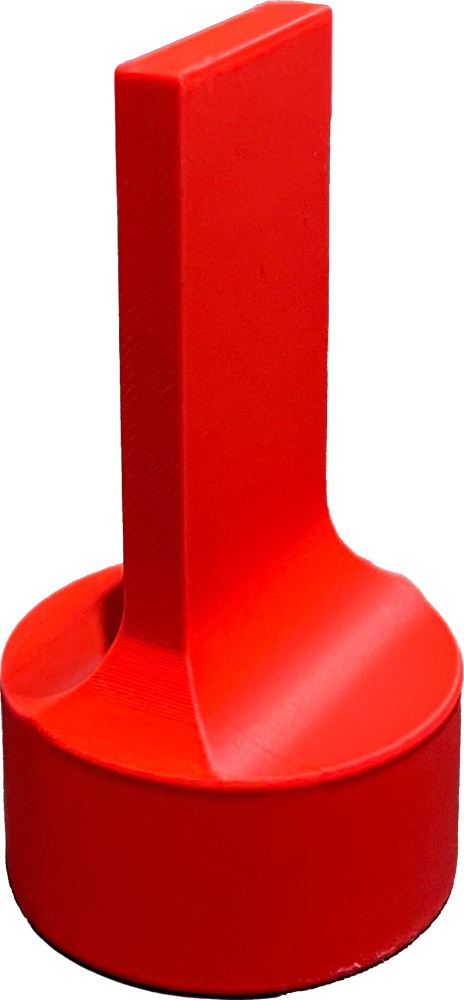}
    & \includegraphics[height=1cm]{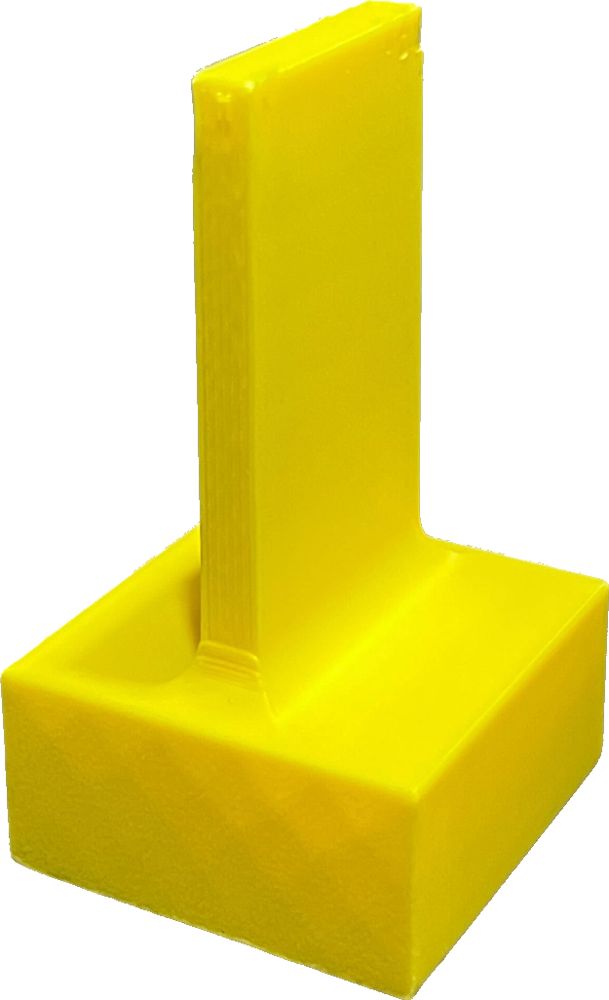}
    & \includegraphics[height=1cm]{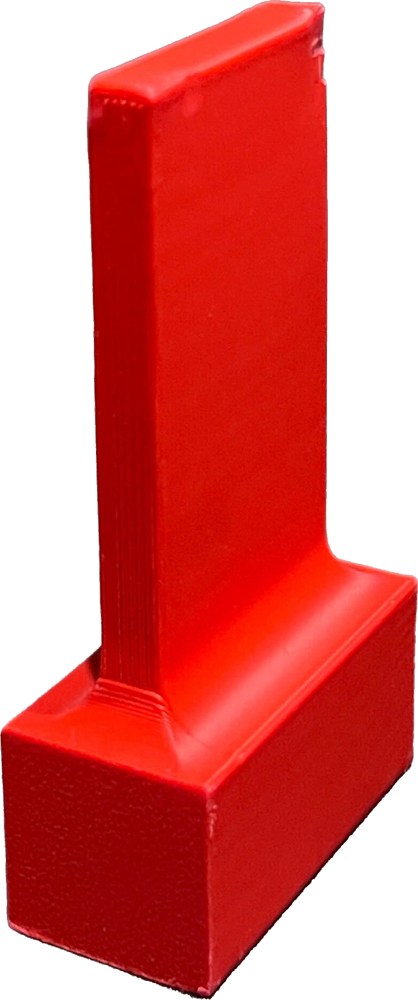}
    & \includegraphics[height=1cm]{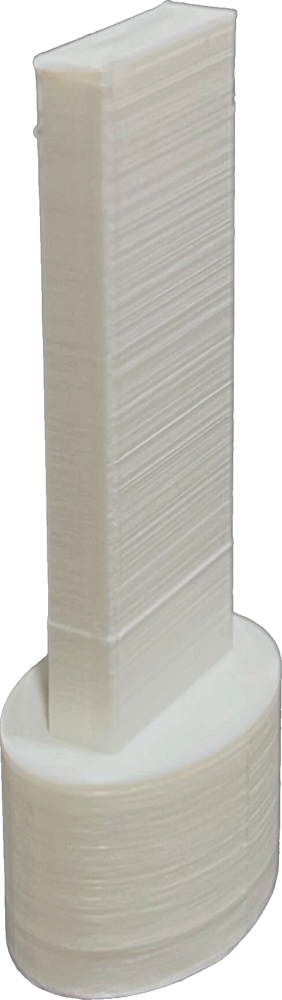}
    & \includegraphics[height=1cm]{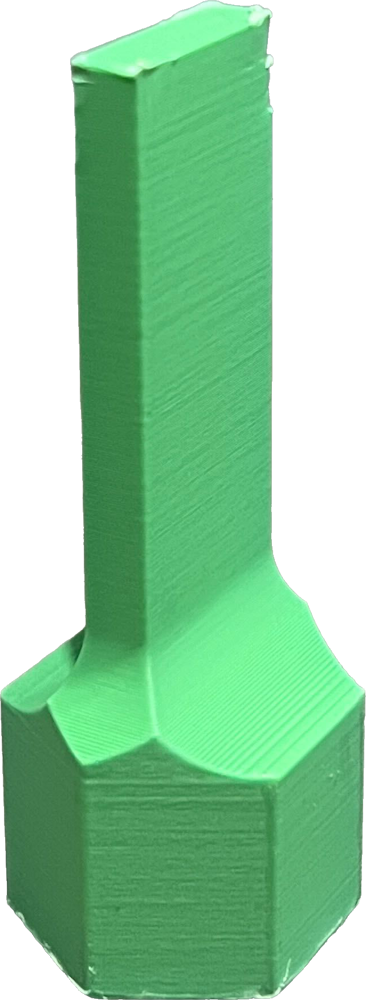}
    & \\
    \midrule
    \quad Target-Only
        & 13/20 & 6/20  & \hcs{9/20}  & 15/20           & 7/20           & 50/100 (50\%) \\
    \quad BC-Prior
        & \hcb{\textbf{19/20}} & \hcs{10/20} & 4/20  & 17/20   & \hcs{14/20}    & \hcs{64/100 (64\%)} \\
    \midrule
    \quad Retrieval-Single~\cite{du2023behavior}
        & 13/20 & 7/20  & 7/20  & 14/20           & \hcb{\textbf{15/20}} & 56/100 (56\%) \\
    \quad Retrieval-Full~\cite{kuroki2024iros}
        & 16/20 & 9/20  & 7/20  & \hcs{19/20} & 13/20        & \hcs{64/100 (64\%)} \\
    \quad Retrieval-Window~\cite{nasiriany2022sailor}
        & 16/20 & \hcs{10/20} & 6/20  & 16/20      & 7/20           & 55/100 (55\%) \\
    \midrule
    \quad \textbf{PHASE (Ours)}
        & \hcs{18/20} & \hcb{\textbf{12/20}} & \hcb{\textbf{13/20}} & \hcb{\textbf{20/20}} & \hcs{14/20} & \hcb{\textbf{77/100 (77\%)}} \\
    \bottomrule
    \end{tabular*}
    ~\\[6pt]
    \parbox{\textwidth}{\scriptsize \textit{Note:}
We report 95\% Wilson confidence intervals for total success rates (n=100):
PHASE 77\% [68\%, 84\%]; BC-Prior 64\% [54\%, 73\%];
Retrieval-Full 64\% [54\%, 73\%];
Retrieval-Single 56\% [46\%, 65\%];
Retrieval-Window 55\% [45\%, 64\%];
Target-Only 50\% [40\%, 60\%].
}
\end{table*}

\textbf{Task and Datasets.}
We evaluate five peg geometries: circle and square, which are included in $\mathcal{D}_{\text{prior}}$, and rectangle, oval, and hexagon, which are absent from $\mathcal{D}_{\text{prior}}$. We refer to these groups as seen and unseen shapes, respectively.
The prior dataset
$\mathcal{D}_{\text{prior}}$ consists of 122
demonstrations, including 61 episodes with square pegs
and 61 with circular pegs, intentionally limited to
two geometries to test whether phase-aware retrieval
can transfer contact strategies from a small prior to
geometrically distinct targets. For each target task, we
collect a small dataset $\mathcal{D}_{\text{target}}$
containing 4 demonstrations via teleoperation using a
VR controller (HTC VIVE Controller, HTC Corporation,
Taiwan). The robot grasps each peg in a vertical
orientation. Sampling and control frequencies are set
to 50~Hz. Each episode begins from one of four initial
poses uniformly distributed around the hole and
terminates upon successful insertion. Episodes last
approximately 150--350 time steps (3--7 seconds). To evaluate robustness under state distribution shift, we
additionally test from starting positions not included
in either $\mathcal{D}_{\text{prior}}$ or
$\mathcal{D}_{\text{target}}$.
For these trials, we randomly perturb the initial peg position by up to $\pm$ 2\,cm in both x and y directions in the horizontal task frame (origin at the hole center). The offsets are applied relative to the nominal teleoperated starting pose.

\textbf{Baselines.} We compare PHASE against
retrieval strategies. Direct comparison with these full systems is
not feasible due to differences in input modalities,
task domains, and hardware. However, retrieval
granularity, whether to retrieve single state-action
pairs, fixed-length segments, or full episodes, is a
transferable design choice independent of these
factors. We therefore extract the core retrieval
strategy from each method and instantiate it within
our shared framework (same ACT policy, multimodal
tactile-proprioceptive inputs, and retrieval budget
$k{=}20$), isolating the effect of retrieval
granularity from system design. Target-Only and BC-Prior
serve as non-retrieval baselines.
\begin{itemize}
\item \textbf{Target-Only:} ACT policy trained solely
on $\mathcal{D}_{\text{target}}$.
\item \textbf{BC-Prior:} ACT policy trained on
$\mathcal{D}_{\text{prior}}$ only, representing a
multi-task policy over all prior geometries.
\item \textbf{Retrieval-Single:} Retrieves
\textit{single} state-action pairs using L2 distance
in embedding space, following the retrieval strategy
of Behavior Retrieval~\cite{du2023behavior}.
\item \textbf{Retrieval-Full:} Retrieves
\textit{complete} episodes using DTW, following
the episode-level retrieval approach
of~\cite{kuroki2024iros}.
\item \textbf{Retrieval-Window:} Retrieves
\textit{fixed-length} segments using sliding windows
of size $w{=}30$ with DTW, following the
segment-based retrieval strategy of
SAILOR~\cite{nasiriany2022sailor}.
\end{itemize}

\textbf{Metrics.}
We report success rate (SR), defined as the proportion
of trials in which the peg is fully inserted into the
hole. Each method is evaluated over 20 trials per
task. A rollout is considered successful if insertion
is completed within a 60-second time limit.

\subsection{Results}
\label{subsec:experiment_results}

\textbf{Comparison with Non-Retrieval Baselines
(\textcolor{blue}{\textbf{RQ1}}).}
We first compare retrieval-augmented methods against
non-retrieval baselines (Target-Only and BC-Prior), which
train a single policy directly on demonstrations
without selecting relevant prior data. As shown in
\Cref{tab:main_results}, PHASE achieves the highest
total success rate (77\%),  
a 13-percentage-point improvement over BC-Prior (64\%) and a 27-percentage-point improvement over Target-Only (50\%).
The relatively weaker performance of BC-Prior suggests
that learning geometry-specific contact strategies
within a single parametric policy is inherently
challenging. Prior work reports that generalist
assembly policies degrade as task diversity increases
due to geometric variation and gradient interference
across tasks~\cite{guo2025srsa}. In our setting,
BC-Prior must encode precise contact transitions for
multiple peg shapes from limited demonstrations.
Retrieval-based conditioning alleviates this burden by
anchoring policy learning to interaction-similar and
phase-consistent segments, reducing cross-task
interference. Note that BC-Prior trains on all 122 prior
demonstrations (${\sim}$39,000 frames), while PHASE augments only 4 target
demonstrations with retrieved segments (15,548
frames). 
Despite being trained on substantially more prior data, BC-Prior underperforms PHASE by 13 percentage points, suggesting that selecting prior experience based on target-relevant phase structure can be more effective than indiscriminately training on the entire prior dataset.

\begin{table}[!t]
\centering
\caption{Offline retrieval characteristics across
retrieval strategies ($k{=}20$). $\downarrow$: lower is better.}
\label{tab:offline}
\begin{tblr}{
  width = \linewidth,
  colspec = {Q[452]Q[104]Q[129]Q[131]Q[115]},
  column{even} = {c},
  column{3} = {c},
  column{5} = {c},
  cell{1}{1} = {font=\bfseries},
  cell{1}{5} = {font=\bfseries},
  hline{1,7} = {-}{0.08em},
  hline{2} = {-}{0.05em},
}
Metric & {\textbf{Single}\\\textbf{Pairs}} & {\textbf{Fixed}\\\textbf{Windows}} & {\textbf{Full}\\\textbf{Episodes}} & PHASE\\
Training frames & 4,479 & 8,820 & 10,596 & 15,548\\
Phase balance (0--1) & 0.99 & 0.98 & 0.87 & 0.98\\
Temporal coherence & 0.72 & 1.00 & 1.00 & 1.00\\
Boundary coverage & 4\% & 78\% & 100\% & 100\%\\
Redundancy ($\downarrow$) & 0.84 & 0.72 & 0.35 & 0.38
\end{tblr}
\end{table}

\begin{table*}[t]
    \centering
    \footnotesize
    \caption{%
        Success rate under unseen starting positions
        (20 trials per shape; 100 total).
        \hcb{Best} and \hcs{second-best} results per
        column are highlighted.}
    \label{tab:state_shift}
    \begin{tabular*}{\textwidth}{@{\extracolsep{\fill}}lcccccc}
    \toprule
    \textbf{Method}
        & \textbf{Circle}
        & \textbf{Square}
        & \textbf{Rectangle}
        & \textbf{Oval}
        & \textbf{Hexagon}
        & \textbf{Total} \\
    \midrule
    \quad Target-Only
        & 0/20  & 0/20  & 0/20  & 0/20  & 0/20  & 0/100 (0\%)   \\
    \quad BC-Prior
        & 5/20  & 2/20  & 3/20  & \hcs{5/20} & \hcs{2/20} & \hcs{17/100 (17\%)} \\
    \midrule
    \quad Retrieval-Single~\cite{du2023behavior}
        & 0/20  & 1/20  & 1/20  & 0/20  & 0/20  & 2/100 (2\%)   \\
    \quad Retrieval-Full~\cite{kuroki2024iros}
        & 4/20  & 4/20  & 4/20  & 2/20  & 0/20  & 14/100 (14\%) \\
    \quad Retrieval-Window~\cite{nasiriany2022sailor}
        & \hcs{6/20} & \hcs{5/20} & \hcs{5/20} & 0/20 & 0/20 & 16/100 (16\%) \\
    \midrule
    \quad \textbf{PHASE (Ours)}
        & \hcb{\textbf{12/20}} & \hcb{\textbf{8/20}} & \hcb{\textbf{12/20}}
        & \hcb{\textbf{9/20}}  & \hcb{\textbf{6/20}} & \hcb{\textbf{47/100 (47\%)}} \\
    \bottomrule
    \end{tabular*}
    ~\\[6pt]
\parbox{\textwidth}{\scriptsize \textit{Note:}
We report 95\% Wilson confidence intervals for total success rates (n=100):
PHASE 47\% [38\%, 57\%]; BC-Prior 17\% [11\%, 26\%];
Retrieval-Window 16\% [10\%, 24\%]; Retrieval-Full 14\% [9\%, 22\%];
Retrieval-Single 2\% [1\%, 7\%]; Target-Only 0\% [0\%, 4\%].
}
\end{table*}

\textbf{Effect of Retrieval Strategy
(\textcolor{blue}{\textbf{RQ2}}).}
We next compare across retrieval-based methods
(Retrieval-Single, Retrieval-Full, Retrieval-Window),
which differ in how they select segments from prior
data. As summarized in \Cref{tab:main_results},
single-pair retrieval achieves 56\% total success,
confirming that isolated state-action matching is
insufficient for insertion tasks. Full-episode
retrieval increases performance to 64\%
(\rev{8 percentage points over single-pair}), demonstrating the importance of temporal context. However, fixed-window retrieval performs comparably to
single-pair retrieval (55\%) despite providing nearly twice the training data, suggesting that rigid
segmentation fails to accommodate the variable duration
of insertion phases and that data volume alone does not
explain the gap. In contrast, PHASE's variable-length
segmentation achieves 77\% total success, outperforming full-episode and fixed-window retrieval by 13 and 22 percentage points, respectively.
These results indicate that
phase-consistent, variable-length retrieval is
substantially more effective than both pointwise and
fixed-length alternatives.
To further understand these differences, we perform
offline analysis to characterize the retrieved training
sets produced by each retrieval
strategy~(\Cref{tab:offline}). Since behavior cloning
directly imitates its training data, the composition
of the retrieved set determines what observation-action
patterns the policy learns to reproduce. We evaluate
five properties: \emph{training volume} (unique frames
available for policy learning after deduplication),
\emph{phase balance} (evenness of search vs.\ insert
coverage), \emph{temporal coherence} (fraction of
retrieved frames forming contiguous sequences),
\emph{boundary coverage} (whether the search-to-insert
transition is captured within $\pm$20 frames), and
\emph{redundancy} (fraction of duplicate retrievals).
Single-pair retrieval yields only 4,479 unique frames
with 84\% redundancy and low boundary coverage
(4\%), meaning the policy rarely observes the critical
contact transition. Fixed-window retrieval ($w{=}30$)
nearly doubles the unique frame count to 8,820 yet
achieves comparable performance, as rigid
windows miss boundaries in 22\% of retrieved episodes
and exhibit high redundancy (0.72). When a fixed window
straddles two contact phases from mismatched
demonstrations, the action chunks sampled from that
region contain conflicting contact dynamics,
introducing noise into the training signal.
Full-episode retrieval guarantees boundary coverage and
temporal coherence but provides fewer effective
training frames (10,596) and the lowest phase balance
(0.87), as entire episodes over-represent the longer
insert phase.
In contrast, PHASE achieves the strongest overall
profile: the most unique training data (15,548 frames),
full boundary coverage (100\%), near-perfect phase
balance (0.98), and low redundancy (0.38\rev{, comparable to full-episode retrieval}). 
Treating each retrieved segment as an independent trajectory prevents action chunks from crossing artificial segment boundaries, while complete target demonstrations retain the genuine search-to-insert transition.
These results suggest that retrieval structure contributes beyond data quantity alone.

\textbf{Generalization Under Distribution Shift
(\textcolor{blue}{\textbf{RQ3}}).}
We evaluate robustness under two forms of shift:
geometry shift (unseen shapes) and state distribution
shift (unseen starting positions). Under geometry
shift, PHASE achieves \rev{approximately} 78\% success on unseen shapes,
improving over BC-Prior (\rev{approximately} 58\%) by 20 percentage points~(\Cref{tab:main_results}). Gains are most
pronounced on rectangle insertion (65\% vs.\ 20\%
\rev{for BC-Prior}).
Ovals achieve 100\%, which we attribute to sharing the smooth contact profile of circles (90\% success), whereas rectangles, which share the edge-alignment demands of squares (60\%), remain more challenging.
Under state distribution shift, the performance gap
widens substantially. 
As shown in
\Cref{tab:state_shift}, Target-Only collapses to 0\%
success, \rev{as the policy effectively memorizes
the four training trajectories and cannot generalize
to novel starting positions}.
Retrieval-based baselines partially
mitigate this, with BC-Prior achieving 17\% and
Retrieval-Full achieving 14\%. In contrast, PHASE
maintains 47\% success across shapes, a 30-percentage-point improvement over BC-Prior.
\rev{We qualitatively observe that phase-aligned
retrieval exposes the policy to diverse search
directions from multiple prior demonstrations,
allowing it to adapt when shifted starting positions
require trajectories not seen in any single
demonstration. For example, if a shifted start
requires sliding right then down, the policy can
compose these movements from separate retrieved
search segments that each cover a different direction.
We hypothesize this compositionality is a key factor
behind the large improvement under distribution shift.}
These results indicate that structuring retrieval
around interaction-defined contact phases is
particularly important when contact timing and phase
transitions deviate from the training distribution.

\textbf{Failure Analysis
(\textcolor{blue}{\textbf{RQ4}}).}
A common failure mode is exceeding the Cartesian force limits of 70 N along the x- and y-axes and 100 N along the z-axis. This triggers a protective stop, typically when downward force is applied before sufficient alignment.
Other failures include an inability to locate the hole within the time limit and unrecovered workspace drift, highlighting limitations in large-offset recovery and contact-state estimation.

\section{Conclusion}
\label{sec:conclusion}
We presented PHASE, a framework for retrieval-augmented policy learning that leverages compliance-enabled tactile and force signals to discover and exploit the phase structure of \rev{insertion tasks}. 
Across five peg geometries, PHASE outperforms the strongest non-phase-aware baseline by 13 percentage points overall and by 30 percentage points under state distribution shift. 
These results \rev{suggest} that aligning retrieval with interaction-defined phase structure is \rev{a useful design principle} for robust \rev{peg-in-hole insertion}.

\textbf{Limitations and Future Work.}
Segmentation currently
relies on \rev{tactile-estimated torque} heuristics tailored
to insertion, which may limit generality. \rev{Replacing
these with learned contact-phase representations from}
multimodal interaction data, for example through
tactile dynamics models trained on play
data~\cite{ai2024robopack}\rev{, could extend PHASE to
broader task families}. Evaluation is restricted
to peg-in-hole insertion with a limited set of
geometries\rev{. Leveraging large-scale open multimodal
robot datasets~\cite{sliwowski2025reassemble} could
allow PHASE to scale across diverse contact-rich
tasks}. Finally, while PHASE improves robustness, it
does not explicitly model contact-state uncertainty
or recovery strategies for large pose deviations.
\rev{Unifying the retrieval representation with the
policy architecture, so that the contact-aware
embeddings used for retrieval also inform policy
decisions, could further improve robustness in
challenging conditions.}

\bibliographystyle{IEEEtran}
\bibliography{draft}

\begin{thebibliography}{10}
\providecommand{\url}[1]{#1}
\csname url@samestyle\endcsname
\providecommand{\newblock}{\relax}
\providecommand{\bibinfo}[2]{#2}
\providecommand{\BIBentrySTDinterwordspacing}{\spaceskip=0pt\relax}
\providecommand{\BIBentryALTinterwordstretchfactor}{4}
\providecommand{\BIBentryALTinterwordspacing}{\spaceskip=\fontdimen2\font plus
\BIBentryALTinterwordstretchfactor\fontdimen3\font minus
  \fontdimen4\font\relax}
\providecommand{\BIBforeignlanguage}[2]{{%
\expandafter\ifx\csname l@#1\endcsname\relax
\typeout{** WARNING: IEEEtran.bst: No hyphenation pattern has been}%
\typeout{** loaded for the language `#1'. Using the pattern for}%
\typeout{** the default language instead.}%
\else
\language=\csname l@#1\endcsname
\fi
#2}}
\providecommand{\BIBdecl}{\relax}
\BIBdecl

\bibitem{tulving2002episodic}
E.~Tulving, ``Episodic memory: From mind to brain,'' \emph{Annual review of
  psychology}, vol.~53, no.~1, pp. 1--25, 2002.

\bibitem{shadmehr1997neural}
R.~Shadmehr and H.~H. Holcomb, ``Neural correlates of motor memory
  consolidation,'' \emph{Science}, vol. 277, no. 5327, pp. 821--825, 1997.

\bibitem{hall2025neural}
S.~Hall-McMaster, M.~S. Tomov, S.~J. Gershman, and N.~W. Schuck, ``Neural
  evidence that humans reuse strategies to solve new tasks,'' \emph{PLoS
  Biology}, vol.~23, no.~6, p. e3003174, 2025.

\bibitem{nasiriany2022sailor}
S.~Nasiriany, T.~Gao, A.~Mandlekar, and Y.~Zhu, ``Learning and retrieval from
  prior data for skill-based imitation learning,'' in \emph{Proc. CoRL}, 2023,
  pp. 2181--2204.

\bibitem{humphreys2022largescale}
P.~C. Humphreys, A.~Guez, O.~Tieleman, L.~Sifre, T.~Weber, and T.~P. Lillicrap,
  ``Large-scale retrieval for reinforcement learning,'' in \emph{Proc.
  NeurIPS}, A.~H. Oh, A.~Agarwal, D.~Belgrave, and K.~Cho, Eds., 2022.

\bibitem{di2024effectiveness}
N.~Di~Palo and E.~Johns, ``On the effectiveness of retrieval, alignment, and
  replay in manipulation,'' \emph{IEEE RA-L}, vol.~9, no.~3, pp. 2032--2039,
  2024.

\bibitem{kuang2025ram}
Y.~Kuang, J.~Ye, H.~Geng, J.~Mao, C.~Deng, L.~Guibas, H.~Wang, and Y.~Wang,
  ``{RAM}: Retrieval-based affordance transfer for generalizable zero-shot
  robotic manipulation,'' in \emph{Proc. CoRL}, 2025, pp. 547--565.

\bibitem{du2023behavior}
M.~Du, S.~Nair, D.~Sadigh, and C.~Finn, ``Behavior retrieval: Few-shot
  imitation learning by querying unlabeled datasets,'' in \emph{Proc. RSS},
  2023.

\bibitem{lin2024flowretrieval}
L.-H. Lin, Y.~Cui, A.~Xie, T.~Hua, and D.~Sadigh, ``{FlowRetrieval}:
  Flow-guided data retrieval for few-shot imitation learning,'' in \emph{Proc.
  CoRL}, 2024, pp. 4084--4099.

\bibitem{strap2025}
M.~Memmel, J.~Berg, B.~Chen, A.~Gupta, and J.~Francis, ``{STRAP}: Robot
  sub-trajectory retrieval for augmented policy learning,'' in \emph{Proc.
  ICLR}, 2025.

\bibitem{belkhale2024rt}
S.~Belkhale, T.~Ding, T.~Xiao, P.~Sermanet, Q.~Vuong, J.~Tompson, Y.~Chebotar,
  D.~Dwibedi, and D.~Sadigh, ``{RT-H}: Action hierarchies using language,'' in
  \emph{Proc. RSS}, 2024.

\bibitem{myers2024policy}
V.~Myers, B.~C. Zheng, O.~Mees, S.~Levine, and K.~Fang, ``Policy adaptation via
  language optimization: Decomposing tasks for few-shot imitation,'' in
  \emph{Proc. CoRL}, 2024, pp. 1402--1426.

\bibitem{zacks2007event}
J.~M. Zacks and K.~M. Swallow, ``Event segmentation,'' \emph{Current Directions
  in Psychological Science}, vol.~16, no.~2, pp. 80--84, 2007.

\bibitem{morgan2023towards}
A.~S. Morgan, Q.~Bateux, M.~Hao, and A.~M. Dollar, ``Towards generalized robot
  assembly through compliance-enabled contact formations,'' in \emph{Proc.
  ICRA}, 2023, pp. 8010--8016.

\bibitem{von2020compact}
F.~von Drigalski, K.~Tanaka, M.~Hamaya, R.~Lee, C.~Nakashima, Y.~Shibata, and
  Y.~Ijiri, ``A compact, cable-driven, activatable soft wrist with six degrees
  of freedom for assembly tasks,'' in \emph{Proc. IROS}, 2020, pp. 8752--8757.

\bibitem{khamis2019novel}
H.~Khamis, B.~Xia, and S.~J. Redmond, ``A novel optical 3d force and
  displacement sensor--towards instrumenting the papillarray tactile sensor,''
  \emph{Sensors and Actuators A: Physical}, vol. 291, pp. 174--187, 2019.

\bibitem{kamijo2026tactile}
T.~Kamijo, M.~Nishimura, N.~Shibasaki, J.~Siburian, C.~C. Beltran-Hernandez,
  and M.~Hamaya, ``Tactile memory with soft robot: Robust object insertion via
  masked encoding and soft wrist,'' \emph{IEEE RA-L}, vol.~11, no.~7, pp.
  7844--7851, 2026.

\bibitem{zhao2023learning}
T.~Z. Zhao, V.~Kumar, S.~Levine, and C.~Finn, ``Learning fine-grained bimanual
  manipulation with low-cost hardware,'' in \emph{Proc. RSS}, 2023.

\bibitem{beltran2020variable}
C.~C. Beltran-Hernandez, D.~Petit, I.~G. Ramirez-Alpizar, and K.~Harada,
  ``Variable compliance control for robotic peg-in-hole assembly: A
  deep-reinforcement-learning approach,'' \emph{Applied Sciences}, vol.~10,
  no.~19, p. 6923, 2020.

\bibitem{hogan1984impedance}
N.~Hogan, ``Impedance control of industrial robots,'' \emph{Robotics and
  Computer-{I}ntegrated Manufacturing}, vol.~1, no.~1, pp. 97--113, 1984.

\bibitem{raibert1981hybrid}
M.~Raibert and J.~Craig, ``Hybrid position/force control of manipulators,''
  \emph{J. of Dynamic Systems, Measurement, and Control}, vol. 103, no.~2, pp.
  126--133, 1981.

\bibitem{ankile2024juicer}
L.~Ankile, A.~Simeonov, I.~Shenfeld, and P.~Agrawal, ``Juicer: Data-efficient
  imitation learning for robotic assembly,'' in \emph{Proc. IROS}, 2024, pp.
  5096--5103.

\bibitem{wang2022adaptive}
Y.~Wang, C.~C. Beltran-Hernandez, W.~Wan, and K.~Harada, ``An adaptive
  imitation learning framework for robotic complex contact-rich insertion
  tasks,'' \emph{Frontiers in Robotics and AI}, vol.~8, p. 777363, 2022.

\bibitem{schoettler2020deep}
G.~Schoettler \emph{et~al.}, ``Deep reinforcement learning for industrial
  insertion tasks with visual inputs and natural rewards,'' in \emph{Proc.
  IROS}, 2020, pp. 5548--5555.

\bibitem{tang2023industreal}
B.~Tang, M.~A. Lin, I.~Akinola, A.~Handa, G.~S. Sukhatme, F.~Ramos, D.~Fox, and
  Y.~Narang, ``Industreal: Transferring contact-rich assembly tasks from
  simulation to reality,'' in \emph{Proc. RSS}, 2023.

\bibitem{guo2025srsa}
Y.~Guo, B.~Tang, I.~Akinola, D.~Fox, A.~Gupta, and Y.~Narang, ``{SRSA}: Skill
  retrieval and adaptation for robotic assembly tasks,'' in \emph{Proc. ICLR},
  2025.

\bibitem{yu2024mimictouch}
K.~Yu, Y.~Han, Q.~Wang, V.~Saxena, D.~Xu, and Y.~Zhao, ``{MimicTouch}:
  Leveraging multi-modal human tactile demonstrations for contact-rich
  manipulation,'' in \emph{Proc. CoRL}, 2024.

\bibitem{pmlr-v229-guzey23a}
I.~Guzey, B.~Evans, S.~Chintala, and L.~Pinto, ``Dexterity from touch:
  Self-supervised pre-training of tactile representations with robotic play,''
  in \emph{Proc. CoRL}, 2023, pp. 3142--3166.

\bibitem{dong2021tactile}
S.~Dong, D.~K. Jha, D.~Romeres, S.~Kim, D.~Nikovski, and A.~Rodriguez,
  ``{Tactile-RL} for insertion: Generalization to objects of unknown
  geometry,'' in \emph{Proc. ICRA}, 2021, pp. 6437--6443.

\bibitem{jin2024visual}
P.~Jin, B.~Huang, W.~W. Lee, T.~Li, and W.~Yang, ``Visual-force-tactile fusion
  for gentle intricate insertion tasks,'' \emph{IEEE RA-L}, vol.~9, no.~5, pp.
  4830--4837, 2024.

\bibitem{romano2011human}
J.~M. Romano, K.~Hsiao, G.~Niemeyer, S.~Chitta, and K.~J. Kuchenbecker,
  ``Human-inspired robotic grasp control with tactile sensing,'' \emph{IEEE
  Transactions on Robotics}, vol.~27, no.~6, pp. 1067--1079, 2011.

\bibitem{jang2025tactile}
H.~Jang, J.~Bae, and K.~Haninger, ``Tactile sensor-based estimation of grasp
  force and contact state with soft fingers,'' \emph{IEEE RA-L}, vol.~10,
  no.~7, pp. 7246--7253, 2025.

\bibitem{lee2022contact}
H.~Lee, S.~Park, K.~Jang, S.~Kim, and J.~Park, ``Contact state estimation for
  peg-in-hole assembly using gaussian mixture model,'' \emph{IEEE RA-L},
  vol.~7, no.~2, pp. 3349--3356, 2022.

\bibitem{zhang2025ta}
Z.~Zhang, H.~Xu, Z.~Yang, C.~Yue, Z.~Lin, H.-a. Gao, Z.~Wang, and H.~Zhao,
  ``{TA-VLA}: Elucidating the design space of torque-aware
  vision-language-action models,'' in \emph{Proc. CoRL}, 2025, pp. 4019--4037.

\bibitem{zhang2022learning}
X.~Zhang, S.~Jin, C.~Wang, X.~Zhu, and M.~Tomizuka, ``Learning insertion
  primitives with discrete-continuous hybrid action space for robotic assembly
  tasks,'' in \emph{Proc. ICRA}, 2022, pp. 9881--9887.

\bibitem{wu2023masked}
P.~Wu, A.~Majumdar, K.~Stone, Y.~Lin, I.~Mordatch, P.~Abbeel, and
  A.~Rajeswaran, ``Masked trajectory models for prediction, representation, and
  control,'' in \emph{Proc. ICML}, 2023, pp. 37\,607--37\,623.

\bibitem{salvador2007toward}
S.~Salvador and P.~Chan, ``Toward accurate dynamic time warping in linear time
  and space,'' \emph{Intelligent data analysis}, vol.~11, no.~5, pp. 561--580,
  2007.

\bibitem{kuroki2024iros}
S.~Kuroki, M.~Nishiura, and T.~Kozuno, ``Multi-agent behavior retrieval:
  Retrieval-augmented policy training for cooperative push manipulation by
  mobile robots,'' in \emph{Proc. IROS}, 2024, pp. 12\,671--12\,678.

\bibitem{ai2024robopack}
B.~Ai, S.~Tian, H.~Shi, Y.~Wang, C.~Tan, Y.~Li, and J.~Wu, ``{RoboPack}:
  Learning tactile-informed dynamics models for dense packing,'' in \emph{ICRA
  Workshop: 3D Visual Representations for Robot Manipulation}, 2024.

\bibitem{sliwowski2025reassemble}
D.~Sliwowski, S.~Jadav, S.~Stanovcic, J.~Orbik, J.~Heidersberger, and D.~Lee,
  ``Reassemble: A multimodal dataset for contact-rich robotic assembly and
  disassembly,'' in \emph{Proc. RSS}, 2025.

\end{thebibliography}

\end{document}